# The efficacy of various machine learning models for multi-class classification of RNA-seq expression data

Sterling Ramroach[1], Melford John[2], and Ajay Joshi[1]

[1] Department of Electrical and Computer Engineering
[2] Department of Pre-Clinical Sciences
The University of the West Indies, St. Augustine campus
sramroach@gmail.com

**Abstract.**
Late diagnosis and high costs are key factors that negatively impact the care of cancer patients worldwide. Although the availability of biological markers for the diagnosis of cancer type is increasing, costs and reliability of tests currently present a barrier to the adoption of their routine use. There is a pressing need for accurate methods that enable early diagnosis and cover a broad range of cancers. The use of machine learning and RNA-seq expression analysis has shown promise in the classification of cancer type. However, research is inconclusive about which type of machine learning models are optimal. The suitability of five algorithms were assessed for the classification of 17 different cancer types. Each algorithm was fine-tuned and trained on the full array of 18,015 genes per sample, for 4,221 samples (75 % of the dataset). They were then tested with 1,408 samples (25 % of the dataset) for which cancer types were withheld to determine the accuracy of prediction. The results show that ensemble algorithms achieve 100% accuracy in the classification of 14 out of 17 types of cancer. The clustering and classification models, while faster than the ensembles, performed poorly due to the high level of noise in the dataset. When the features were reduced to a list of 20 genes, the ensemble algorithms maintained an accuracy above 95% as opposed to the clustering and classification models.

## 1 Introduction

Recent technological advances in molecular biology have resulted in cheaper and easier methods to perform RNA-seq expression analysis to extract the gene expression profile of a cell or tissue sample. These advances lead to the creation of large datasets of gene expression profiles for various diseases. Given these datasets, new fields emerged within bioinformatics to include diagnosis of diseases using gene expression data and the prediction of clinical outcomes with respect to treatment. RNA-seq expression analysis provides a quantitative measure of the expression levels of genes in a cell. Current research estimates the existence of 19,000 genes per cell [1]. This data has been shown to contain the information necessary to diagnose cancer type.

Given the vast amount of data, it is not feasible for humans to understand relationships between samples and genes without assistance from a machine. Machine learning and other types of artificial intelligence have been used to build predictive models to classify and understand the relationships between gene expression levels and cancer type [2-8]. Previous works report wide variations in either accuracy, error rates, kappa coefficient, specificity, sensitivity, and so on. Prior studies also performed limited experiments using a small number of classes, classifiers, varying number of genes, small datasets, and a limited number of cancer types. A wealth of data arising out of expansive cancer genome projects such as The Cancer Genome Atlas (TCGA) now exists, which can be used to train machine learning algorithms to build decision models to diagnose cancer type, and possibly to identify causative genetic factors. A thorough understanding of the strengths and weaknesses of modern classifiers is necessary in order to build a successful RNA-seq expression diagnostic model. This study aims to provide sufficient information, to highlight the best classifiers for this type of data as well as similarly structured datasets, and to illustrate a structured comparison of the performance of various machine learning algorithms.

Podolsky et al. classified five cancer types using various gene expression datasets and found that K-nearest neighbor outperformed the support vector machine and an ensemble classifier [9]. Tarek et al. also found that the K-nearest neighbor achieves higher accuracy than ensemble methods for datasets of three types of cancer [10]. Azar et al. [3] and Uriarte et al. [11] showed that the Random Forest algorithm is preferable for gene selection using gene expression data. Al-Rajab et al. [1] and Tan and Gilbert [12] also highlighted the potential of using ensemble machine learning algorithms on gene expression data.

In this study, using a much larger number of tumours, a wider range of cancer types, and a larger number of genes, the possibility of achieving near 100% accuracy in the diagnosis of cancer type is investigated along with an investigation into the performances of the various models. The Random Forest (RF), along with other ensemble machine learning algorithms such as Gradient Boosting Machine (GBM), and Random Ferns (RFERN) are further analyzed in this work. The performance of these algorithms were compared with a classification and a clustering algorithm; Support Vector Machine (SVM) and K-Nearest Neighbor (KNN), respectively. Preliminary insight is also provided regarding the most important features or the genes used by the most successful models in relation to cancer type.

## 2. Methods

The dataset downloaded from the Catalogue of Somatic Mutations in Cancer (COSMIC) contained 5,629 samples with the expression levels of 18,019 genes each, resulting in an $n \times m$ matrix, where $n = 5,629$ and $m = 18,019$. The COSMIC data portal lists the gene expression dataset with the name CosmicCompleteGeneExpression.tsv. The version of the dataset downloaded is v80, with 101,406,435 rows in the format illustrated in Figure 1.

| Sample_id | Gene_name | Z_score |
|---|---|---|
| Sample_1 | Gene_1 | 2.41 |
| Sample_1 | Gene_2 | 1.21 |
| Sample_2 | Gene_1 | 0.97 |
| Sample_1 | Gene_3 | -0.20 |
| ... | ... | ... |
| Sample_$n$ | Gene_$m$ | -1.20 |

**Fig. 1.** Format of the cosmic_complete_gene_expression dataset from the COSMIC repository.

Inferring meaning from this raw dataset is difficult as the samples are not classified. Pre-processing is required before any machine learning models can be built. COSMIC also provides another dataset titled CosmicSample.tsv, which contains data on the primary site and the histology subtype for tumour samples as shown in Figure 2.

| Sample_id | Primary_site | Histology_Subtype |
|---|---|---|
| Sample_1 | Lung | Adenocarcinoma |
| Sample_2 | Lung | Squamous cell Carcinoma |
| Sample_3 | Prostate | Adenocarcinoma |
| ... | ... | ... |
| Sample_$n$ | primary site | histology subtype |

**Fig. 2.** Layout of the cosmic_sample dataset displaying the primary site and histology subtype of each sample.

Supervised machine learning models require the data to be labelled and thus, a combination of both datasets is needed for further analysis. The format of the processed dataset used in these experiments is illustrated in Figure 3. There are 17 classes (types of cancer) in the dataset, ranging from 48 to 601 rows (samples).

Information on how to download the data is provided on the COSMIC website located at URL http://cancer.sanger.ac.uk/cosmic/download. All gene expression data provided by COSMIC were obtained from The Cancer Genome Atlas portal, which provides strict guidelines on the preparation of tumour samples. According to The Cancer Genome Atlas, gene expression (mRNA) data for all tumours were generated by the University of North Carolina at Chapel Hill, Chapel Hill, N.C and all expression data are presented in a normalized form as z-scores [11]. All experiments were performed in the R environment [13] on an Asus Republic of Gamers G75V laptop with an Intel Core i7 2.4 GHz processor and 16 GB RAM operating on Microsoft Windows 8.1 64-bit operating system.

|  | Sample_id | Primary_site | Histology_subtype | Gene_1 | Gene_2 | ... | Gene_m |
|---|---|---|---|---|---|---|---|
| | Sample_1 | Lung | Adenocarcinoma | 2.41 | 1.21 | ... | 0.99 |
| | Sample_2 | Lung | Squamous cell Carcinoma | 0.97 | 1.21 | ... | 2.33 |
| $n$ samples | Sample_3 | Prostate | Adenicarcinoma | 2.39 | 2.05 | ... | -1.20 |
| | ... | ... | ... | ... | ... | ... | -0.55 |
| | Sample_$n$ | primary site | histology subtype | 1.23 | 1.21 | ... | -1.20 |

(columns Gene_1 ... Gene_m are $m$ genes)

**Fig. 3.** Matrix representation of the pre-processed RNA-seq expression analysis, where rows represent samples, and columns represent the z-scores of each gene.

RF builds classification trees are using a bootstrap sample of the dataset [6, 14, 15]. Each split is derived by searching a random subset (chosen by varying split points) of the given variables (16,718 genes) as the candidate set [16]. Although memory intensive, all trees are grown using all of the features in the training set to attain low-bias trees. Many deep uncorrelated trees are also grown to ensure low variance. This reduced bias and variance results in a low error rate. When classifying a sample, all trees in the forest output a vote declaring whether or not the new sample belongs to its class. The random forest classifies the new sample using the highest voted class. This ensemble is not prone to over-fitting since splitting points are randomly chosen (there will always be a random distribution). Another ensemble algorithm is the GBM. GBM is a nonparametric machine learning approach for classification based on sequentially adding weak learners to the ensemble [15, 17]. GBM reduces the error in its model by repeatedly combining weak predictors. One new weak learner is added to the ensemble with the sole purpose of complementing the existing weak learners in the model. If the new learner does not complement the ensemble, it is discarded. This process accounts for the usually slow training process of a GBM.

RFERN can be considered a constrained decision tree ensemble initially created for image processing tasks [18]. It is somewhat similar to a binary decision tree where all splitting criteria are identical making it a semi-naïve bayes classifier. The ensemble is built by randomly choosing different subsets of features for each fern. The SVM used a one-versus-rest technique where each class was separated into groups whereby it is positive compared to all other classes [19-21]. It finds hyper-planes that maximally separates classes. Due to the high dimensionality of the data, attributes are projected onto a high dimensional plane which make the data less likely to be linearly separable. This technique is known to perform poorly on high dimensional data. The KNN algorithm plots all attributes as points in a complex dimensional space (the number of genes). Using Manhattan distance, this algorithm classifies a new sample based on the votes of the nearest neighbors [21, 22]. KNN also performs poorly when tasked with classifying high dimensional data. Initial clusters fail to adapt to the training data. Attributes which were clustered incorrectly cannot be relocated at the end of modelling. Feature selections methods are commonly paired with SVM and KNN to avoid these issues.

## 3. Results and Discussion

The full dataset contained 5,629 samples comprised of 17 classes. The classes were unevenly distributed with the smallest class consisting of 48 samples, compared to the largest class of 601 samples. There were also NA, or NULL values for some attributes (genes) of various classes. The full dataset contained a total of 18,015 attributes, however, these were filtered to remove any NA or NULL values, resulting in 16,718 attributes per sample in the training set. After the models were built, they were then assessed with the task of classifying all samples in the test set, with the full 18,015 genes. Previous works focused on reducing the number of genes in training the prediction models [8, 11, 23, 24]. Feature reduction is performed after analysis of the results of utilizing the entire genome (or a larger number of genes than previously studied) when building prediction models. The dataset has the peculiar characteristic of having the number of attributes, orders of magnitude higher than the number of samples. High dimensional data often contains a high level of noise which is evident in the performance of some models.

Initially, each machine learning algorithm built its prediction model by examining a training set of gene expression data for which all 17 cancer types were visible. Models were built by finding relationships between the levels of expression of subsets of genes to a cancer type. The models were then assessed with a test set of gene expression data for which cancer types were withheld. Each algorithm was fine-tuned and trained on the full array of 18,015 genes per sample, for 4,221 samples (75 % of the dataset). They were then tested with 1,408 samples (25 % of the dataset). Table 1 shows the training and testing times for all models.

**Table 1.** Training and testing times of all models.

|  | RF | GBM | RFERN | SVM | KNN |
|---|---|---|---|---|---|
| **Average training time (s)** | 15,859 | 21,445 | 1,332 | 3,271 | 1,524 |
| **Testing time (s)** | 441 | 294 | 338 | 749 | 411 |

Table 2 decomposes the overall classification accuracy by providing the accuracy for each class in the test set. The time taken for RF to be trained is 15,859 seconds. Although this is the second slowest algorithm, it attained the highest accuracy (99.89%) as shown in Table 2. The time taken to meticulously sift through the noise in the dataset lead to an overall better model. GBM accurately classified 99.68% of samples in the test set. These results imply that greedy boosting can be a suitable direction in modelling gene expression data. RFERN classified 94.12% of test samples. It achieved the third best accuracy with the fastest training time of 1,332 seconds. This model was trained faster than RF and GBM because training time grows linearly with fern size, rather than exponentially with tree depth (as seen with the other ensemble techniques).

There is a strong correlation between training times and prediction accuracy. The ensemble machine learning methods are all trained differently, with GBM and RF requiring the most time. Whereas the classification and clustering algorithms built their models relatively quickly. Ensemble algorithms outperformed SVM and KNN due to the nature of the problem. The number of genes per sample is much greater than the number of samples. There are 16,718 genes per sample, however, not all of these genes help differentiate between various cancer types. Ensemble algorithms are built to sift through the noise in large datasets to extract the core features.

**Table 2.** Accuracy of all models.

| Primary Site | RF | GBM | RFERN | SVM | KNN |
|---|---|---|---|---|---|
| **Histology Subtype** | Accuracy (%) | | | | |
| Central Nervous System Astrocytoma Grade IV | 100 | 100 | 100 | 16.67 | 64.29 |
| Cervix Squamous cell Carcinoma | 100 | 100 | 96.09 | 37.34 | 44.58 |
| Endometrium | 100 | 100 | 98.25 | 0.00 | 5.56 |
| Carcinosarcoma Malignant Mesodermal Mixed Tumour | | | | | |
| Endometrium Endometrioid Carcinoma | 100 | 100 | 80.92 | 72.86 | 78.57 |
| Haematopoietic and Lymphoid Tissue Acute Myeloid Leukaemia | 100 | 100 | 100 | 43.75 | 97.92 |
| Haematopoietic and Lymphoid Tissue Diffuse Large B cell Lymphoma | 97.92 | 97.92 | 95.83 | 0.00 | 45.45 |
| Kidney Chromophobe Renal cell Carcinoma | 100 | 100 | 100 | 0.00 | 86.96 |
| Kidney Clear cell Renal cell Carcinoma | 100 | 100 | 99.81 | 80.80 | 98.40 |
| Large Intestine Adenocarcinoma | 100 | 99.17 | 81.53 | 73.83 | 70.47 |
| Liver Hepatocellular Carcinoma | 100 | 100 | 100 | 54.12 | 95.29 |
| Lung | 100 | 100 | 96.13 | 69.57 | 78.26 |
| Adenocarcinoma | | | | | |
| Lung Squamous cell Carcinoma | 99.80 | 100 | 95.02 | 76.03 | 76.03 |
| Ovary Serous Carcinoma | 100 | 100 | 99.25 | 95.59 | 13.24 |
| Pancreas Ductal Carcinoma | 100 | 100 | 92.26 | 9.38 | 81.25 |
| Prostate Adenocarcinoma | 100 | 100 | 89.96 | 82.79 | 95.90 |
| Stomach Adenocarcinoma | 98.60 | 95.79 | 82.11 | 19.40 | 55.22 |
| Upper Aerodigestive Tract Squamous cell Carcinoma | 100 | 100 | 92.91 | 69.85 | 86.02 |
| **Average** | **99.89** | **99.68** | **94.12** | **47.18** | **75.43** |

The tumours belonging to the classes (cancer primary sites and histology subtypes): Central Nervous System Astrocytoma Grade IV, Cervix Squamous cell Carcinoma, Endometrium Carcinosarcoma Malignant Mesodermal Mixed tumour, Endometrium Endometrioid Carcinoma, Haematopoietic and Lymphoid Tissue Acute Myeloid Leukaemia, Kidney Chromophobe Renal cell Carcinoma, Kidney Clear cell Renal cell Carcinoma, Large Intestine Adenocarcinoma, Liver Hepatocellular Carcinoma, Lung Adenocarcinoma, Lung Squamous cell Carcinoma, Ovary Serous Carcinoma, Pancreas Ductal Carcinoma, Prostate Adenocarcinoma, and Upper Aerodigestive Tract Squamous cell Carcinoma were all classified with 100% accuracy by one of the ensemble algorithms. This implies that the genetic mutations which instigate the formation of these types of tumours share a similar distribution throughout all samples.

Given the stellar performances of RF and GBM, features were reduced based on a combination of the important variables selected by both ensembles. The top 80 genes were extracted and these were used to train the models again. This process was repeated for 60, 40, 20, and finally, 10 genes. Figure 4 illustrates the classification accuracy of all models built on these subsets of genes.

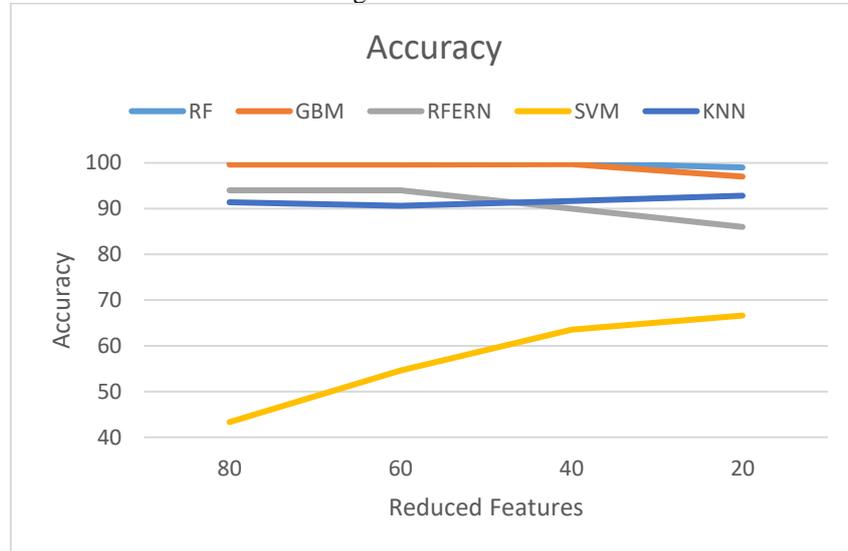

**Fig. 4.** Model accuracy based on a reduction of features.

The accuracy dips when the feature count is 20. These 20 genes were extracted and displayed in Table 3. Defensin Beta, Interferon, and Keratin Associated Proteins are known to be strong cancer driver genes [25, 26]. Actin and Ribonuclease has been shown to play a significant role in some cancer types [27]. Further work is needed to determine the impact of olfactory genes on cancer as the current research is conflicted [28, 29].

**Table 3.** RF-GBM feature selection output.

| Name | Description |
|---:|---|
| ACTL9 | Actin Like 9 |
| ACTRT2 | Actin Related Protein T2 |
| C10orf122 | Testis |
| C17orf105 | Chromosome 17 |
| CLRN2 | Clarin |
| DEFB104A | Defensin Beta |
| DEFB112 | Defensin Beta |
| DEFB119 | Defensin Beta |
| DEFB136 | Defensin Beta |
| HS3ST3B1 | Heparan Sulfate-Glucosamine |
| IFNA7 | Interferon |
| KRTAP10.8 | Keratin Associated Protein |
| KRTAP19.5 | Keratin Associated Protein |
| KRTAP19.6 | Keratin Associated Protein |
| KRTAP24.1 | Keratin Associated Protein |
| OR10G9 | Olfactory Receptor Family |
| OR2M2 | Olfactory Receptor Family |
| OR52E4 | Olfactory Receptor Family |
| OR5A2 | Olfactory Receptor Family |

| RNASE12 | Ribonuclease |

Many tests and statistics were analyzed to ensure integrity of the results. The high levels of accuracy achieved are unlikely to be due to batch bias. The Cancer Genome Atlas gene expression values were generated by laboratory analysis at a single source, therefore errors in the equipment which were used for RNA-seq expression analysis would have been consistent throughout the dataset. Also, all expression values were normalized by conversion to z-scores. There is one major limitation to this study which arises from the data. There are no healthy samples in the dataset. Although the ensemble algorithms performed well at identifying cancer type, we are still unsure of their ability to identify healthy or non-cancerous tumours.

## 4. Conclusion

The use of machine learning in cancer diagnosis is becoming more feasible as algorithms become less prone to error and noise, and as the volume of training data increases. Using machine learning and data from RNA-seq expression, the potential exists for faster and more accurate diagnosis of cancer type. When trained with data derived from RNA-seq expression analysis, the random forest and gradient boosting machine were able to classify cancer type with an accuracy significantly greater than that of a support vector machine and K-nearest neighbors. The random forest and the gradient boosting machine were capable of diagnosing 17 types of cancer with accuracies of 99.89% and 99.68%, respectively. The random ferns algorithm was the third best performing algorithm with an accuracy of 94.12%. It was also the fastest algorithm. The support vector machine and the K-nearest neighbor algorithms classified cancer type with 47.18% and 75.13% accuracy, respectively. This difference in performance is attributed to the given task and the features of the dataset. RNA-seq expression produces a quantitative description of the levels of expression of genes in a cell. There are 18,015 genes and 5,629 samples in the dataset. The number of features (genes) is orders of magnitude higher than the number of samples, hence there is a high level of noise through which these algorithms need to sift. Pairing feature selection with all models lead to a significant improvement in the accuracy of the support vector machine, and K-nearest neighbors to 71.52% and 94.74%, respectively. Defensin beta, keratin associated protein, and the olfactory receptor family were found to be highly influential in the classification of cancer type.